\documentclass[10pt,conference]{IEEEtran}

\usepackage{amsmath,amssymb,amsfonts}
\usepackage{algorithmic}
\usepackage[linesnumbered,ruled,vlined]{algorithm2e}
\usepackage{graphicx}
\usepackage{textcomp}
\usepackage{booktabs}
\usepackage{multirow, multicol}
\usepackage{makecell}
\usepackage{pifont} % for cmark and xmark
\usepackage{adjustbox}
\usepackage[normalem]{ulem}
\usepackage{flushend}
\usepackage{xcolor}
\usepackage[linesnumbered,ruled,vlined]{algorithm2e}

\SetCommentSty{mycommfont}
\usepackage{comment}

\usepackage{tikz}
\usetikzlibrary{shapes.geometric, positioning}
\usepackage{pgfplots}
\pgfplotsset{compat=1.18} % address warning for pgfplots
\usepgfplotslibrary{groupplots}

\definecolor{darkgreen}{rgb}{0, 0.5, 0}
\definecolor{darkred}{rgb}{0.6, 0, 0}
\definecolor{lightblue}{rgb}{0.7, 0.7, 0.9}
\definecolor{lightred}{rgb}{1, 0.75, 0.75}
\newcommand{\cmark}{\textcolor{darkgreen}{\ding{51}}}
\newcommand{\xmark}{\textcolor{darkred}{\ding{55}}}

\newcommand{\Fairify}{\textsf{Fairify}}
\newcommand{\FairQuant}{\textsf{FairQuant}}

\usepackage[scaled=0.9]{helvet} % comment out this instruction to suspend scaling
\newtheorem{definition}{Definition}
\newtheorem{theorem}{Theorem}

\begin{document}

\title{FairQuant: Certifying and Quantifying Fairness of Deep Neural Networks}
\author{
    \IEEEauthorblockN{
    Brian Hyeongseok Kim}
    \IEEEauthorblockA{
%    Department of Computer Science \\
    University of Southern California \\
    Los Angeles, USA}
\and
    \IEEEauthorblockN{
    Jingbo Wang}
    \IEEEauthorblockA{
%    Department of Computer Science \\
    Purdue University\\
    West Lafayette, USA}
\and
    \IEEEauthorblockN{
    Chao Wang}
    \IEEEauthorblockA{
%    Department of Computer Science \\
    University of Southern California \\
    Los Angeles, USA}
}

\maketitle

\begin{abstract}
We propose a method for formally certifying and quantifying \emph{individual fairness} of deep neural networks (DNN). Individual fairness guarantees that any two individuals who are identical except for a legally protected attribute (e.g., gender or race) receive the same treatment. While there are existing techniques that provide such a guarantee, they tend to suffer from lack of scalability or accuracy as the size and input dimension of the DNN increase. 
Our method overcomes this limitation by applying abstraction to a symbolic interval based analysis of the DNN followed by iterative refinement guided by the fairness property. Furthermore, our method lifts the symbolic interval based analysis from conventional \textit{qualitative} certification to \textit{quantitative} certification, by computing the percentage of individuals whose classification outputs are provably fair, instead of merely deciding if the DNN is fair.  
We have implemented our method and evaluated it on deep neural networks trained on four popular fairness research datasets. The experimental results show that our method is not only more accurate than state-of-the-art techniques but also several orders-of-magnitude faster. 
\end{abstract}

\section{Introduction}

The problem of certifying the fairness of machine learning models is more important than ever due to strong interest in applying machine learning to automated decision making in various fields from banking~\cite{castelnovo2020} and healthcare~\cite{paulus2020predictably} to public policy~\cite{rodolfa2021empirical} and criminal justice~\cite{wang2023pursuit}. Since the decisions are  socially sensitive, it is important to certify that the machine learning model indeed treats individuals or groups of individuals fairly. 
However, this is challenging when the model is a deep neural network (DNN) with a large number of hidden parameters and complex nonlinear activations. 
The challenge is also exacerbated as the network size and input dimension increase.
In this work, we aim to overcome the challenge by leveraging abstract interpretation techniques to certify fairness both qualitatively and quantitatively.

Our work focuses on \emph{individual fairness} which, at a high level, requires that similar individuals are treated similarly~\cite{dwork2012fairness}.
Here, \emph{similar individuals} are those who differ only in some legally \emph{protected} input attribute (e.g., gender or race) but agree in the \emph{unprotected} attributes,
and \emph{being treated similarly} means that the DNN generates the same classification output.\footnote{
This notion can be understood as causal fairness~\cite{galhotra_fairness_2017} or dependency fairness~\cite{urban_perfectly_2020}, which is a non-probabilistic form of counterfactual fairness~\cite{kusner2017counterfactual}.
} 
Let the DNN be a function $f: X\rightarrow Y$ from input domain $X$ to output range $Y$, where an individual $x\in X$ is an input and a class label $y\in Y$ is the output.
Assume that each input $x=\langle x_1,\ldots,x_D\rangle$ is a $D$-dimensional vector, and  $x_j$, where $1\leq j \leq D$, is a \emph{protected} attribute.
We say that the DNN is \emph{provably fair} (certified) for the entire input domain $X$ if $f(x)=f(x')$ holds for any two individuals $x \in X$ and $x' \in X$ that differ only in $x_j$ but agree in the unprotected attributes ($\forall x_i$ where $i\neq j$).
Conversely, the DNN is \emph{provably unfair} (falsified) for input domain $X$ if $f(x) \neq f(x')$ holds for any two individuals ($x \in X$ and $x'\in X$) that differ only in the protected attribute.
If the DNN is neither certified nor falsified, it remains \emph{undecided}.

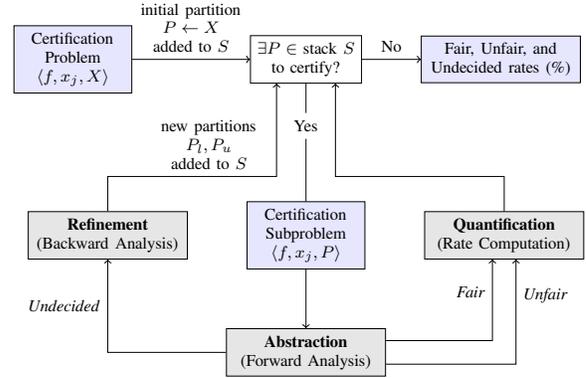
\begin{figure}
\centering
\resizebox{0.9\linewidth}{!}{
\tikzset{
    every node/.style={
    font=\footnotesize,
    align=center
    }
}

\begin{tikzpicture}

\node [draw, rectangle, fill=blue!10, minimum width=2cm]
    (prob) {Certification \\ Problem \\ $\langle f, x_j, X \rangle$};
\node [draw, rectangle, right=2cm of prob]
    (loop) {$\exists P \in$ stack $S$\\ to certify?};
\node [draw, rectangle, right=1cm of loop, fill=blue!10, minimum width=2.7cm]
    (output) {Fair, Unfair, and \\ Undecided rates (\%)};

\draw[->] (prob.east) -- (loop.west) node[midway, above] {initial partition \\ $P \leftarrow X$ \\ added to $S$};
\draw[->] (loop.east) -- (output.west) node[midway, above] {No};

\node [below=0.5cm of loop]
    (yes) {Yes};

\node [draw, rectangle, fill=blue!10, below=2cm of loop, minimum width=2cm]
    (subprob) {Certification \\ Subproblem \\ $\langle f, x_j, P \rangle$};

\node [draw, rectangle, fill=gray!20, below=of subprob, minimum width=2.7cm]
    (forward) {\textbf{Abstraction} \\ (Forward Analysis)};

\node [draw, rectangle, fill=gray!20, left=of subprob, minimum width=2.7cm]
    (backward) { \textbf{Refinement} \\ (Backward Analysis)};

\node [draw, rectangle, fill=gray!20, right=of subprob, minimum width=2.7cm]
    (rates) { \textbf{Quantification} \\ (Rate Computation)};

% Arrows

\draw[->] (loop) -- (yes) -- (subprob) -- (forward);

\draw[->] (forward) -| (backward) node[near end, left] {\textit{Undecided}};
\draw[->] ([yshift=2mm] forward.east) -| ([xshift=-2mm] rates.south) node[pos=0.8, left] {\textit{Fair}};
\draw[->] ([yshift=-2mm] forward.east) -| ([xshift=2mm] rates.south) node[pos=0.83, right] {\textit{Unfair}};

\draw[->] (rates) -- +(0, 1) -| ([xshift=5mm] loop.south);
\draw[->] (backward) -- +(0, 1) -| ([xshift=-5mm] loop.south)
    node[pos=0.3, above] {new partitions \\ $P_l, P_u$ \\ added to $S$};

\end{tikzpicture}
}
\caption{\FairQuant{}  for certifying and quantifying fairness of a DNN  model $f$ where $x_j$ is a protected attribute and $X$ is the input domain.}
\label{fig:overview}
\end{figure}

%There is a difference between \emph{qualitative} certification and \emph{quantitative} certification.
%
Given a DNN $f$, a protected attribute $x_j$, and an input domain $X$, a \emph{qualitative} certification procedure aims to determine whether $f$ is \emph{fair}, \emph{unfair}, or \emph{undecided} for all $x \in X$.  
Qualitative analysis is practically important because, if $f$ is provably fair, the model may be used \emph{as is}, but if $f$ is provably unfair, the model should not be used to make decision for any $x\in X$. 
When the result of qualitative analysis is \emph{undecided}, however, there is a need for \textit{quantitative} analysis, to compute the degree of fairness.  For example, the degree of fairness may be measured by the percentage of individuals in input domain $X$ whose classification outputs  are provably fair.

Both \emph{qualitative} certification and \emph{quantitative} certification are hard problems for deep neural networks. 
While there are many verification tools for deep neural networks, existing verifiers such as ReluVal~\cite{wang_formal_2018}, DeepPoly~\cite{singh_abstract_2019}, and $\alpha$-$\beta$-CROWN~\cite{wang2021beta} focus on certifying perturbation robustness, which is a fundamentally different property, and cannot certify individual fairness. 
%In particular, we will shown in Section 2 that $\alpha$-$\beta$ CROWN is not effective in certifying individual fairness.
%
To the best of our knowledge, the only existing technique for certifying individual fairness of a DNN is \Fairify{}~\cite{biswas_fairify_2023}. However, since it directly analyzes the behavior of a DNN in the concrete domain using the SMT solver, the computational cost is extremely high; as a result, \Fairify{} can only certify tiny networks.  
Furthermore, it cannot quantify the degree of fairness.
Prior works on quantitative analysis of fairness focus on either testing~\cite{angell_themis_2018, zhang_ADF_2020, monjezi_dice_2023}, which do not lead to sound certification, or statistical parity~\cite{albarghouthi2017fairsquare,bastani2019probabilistic} that concerns another type of fairness,  group fairness, which differs significantly from individual fairness. 
%Our work advances the current state of the art in \textit{quantitative fairness verification of neural networks}.

To fill the gap, we propose the first scalable method for \emph{certifying} and \emph{quantifying} individual fairness of a DNN. 
Our method, named \FairQuant{}, takes a certification problem (consisting of the DNN $f$, protected attribute $x_j$, and input domain $X$) as input and returns one of the following three outputs:
(1) \textit{certified (fair)} for all input $x\in X$; 
(2) \textit{falsified (unfair)} for all input $x\in X$;
or (3) \textit{undecided}, meaning that $f$ is neither 100\% \textit{fair} nor 100\% \textit{unfair}.
In the third case, our method also computes the percentage of inputs in $X$ whose classification outputs are provably \emph{fair}.
More specifically, our method provides a \emph{lower bound} of the certified percentage, which can guarantee that the DNN meets a certain requirement, e.g., the DNN is individually fair for at least 80\% of all inputs in $X$.

As shown in Fig.~\ref{fig:overview},  \FairQuant{} iterates through three steps: abstraction (forward analysis), refinement (backward analysis), and quantification (rate computation). 
Assuming that the legally protected attribute $x_j$ has two possible values (e.g., male and female), forward analysis tries to prove that, for each $x$ in the input partition $P$ (which is the entire input domain $X$ initially), flipping the value of the protected attribute of $x$ does not change the model's output.
This is accomplished by propagating two symbolic input intervals $I$ ($\forall x\in P$ that are male) and $I'$ ($\forall x'\in P$ that are female) to compute the two corresponding output intervals that are overapproximated. 
If the classification labels (for all $x$ and $x'$) are the same, our method returns \emph{certified (fair)}. 
On the other hand, if the classification labels (for all $x$ and $x'$) are different, our method returns \emph{falsified (unfair)}. 
In these two cases, 100\% of the inputs in the partition $P$ are resolved. 
 
%corresponding to the size of this region.
%\jwnote{How do you calculate this rate in the forward pass and do you also need to split the input region in the forward pass? } \bknote{Calculating the rate (shown later) is simply calculating the number of (x, x') pairs represented by a given input interval range, and dividing it by the number of all possible (x, x') pairs in the entire domain. Here we assume that there is a finite number of possible (x, x') pairs, albeit very large (up to trillions), since the input vector is not normalized and takes in integers.}

Otherwise,  we perform refinement (backward analysis) by splitting $P$ into partitions $P_l$ and $P_u$ and apply forward analysis to each of these new partitions. 
Since smaller partitions often lead to smaller approximation errors, refinement has the potential to increase the number of certified (or falsified) inputs and decrease in the number of undecided inputs.
%
%Our method for splitting the partition $P$ is to identify an input attribute $x_i$ (where $i\neq j$) and then split its interval from $x_i\in [lb,ub]$ into $x_i\in [lb, (lb+ub)/2]$ for $P_l$ and $x_i\in [(lb+ub)/2,ub]$ for $P_u$.  Toward this end, we propose a refinement strategy based on the gradients computed by the forward pass for both $I$ ($\forall x\in P$) and $I'$ ($\forall x'\in P$).
%
%
%In theory, the iterative refinement based interval analysis is guaranteed to terminate for DNNs that use ReLU activation~\cite{wang_formal_2018}.  However, in practice, the number of refinement iterations may be extremely large for DNNs with high-dimensional inputs.
%
To ensure that our method terminates quickly, we propose two \emph{early termination} conditions based on the refinement depth of each partition $P\subseteq X$.  The refinement depth is the number of times $X$ is partitioned to produce $P$.
There are two predefined thresholds.
%
%\begin{itemize}
%\item 
%
Once the refinement depth exceeds the higher threshold, we classify the partition $P$ as \textit{undecided} and avoid splitting it further.
%\item
%
But if the refinement depth exceeds the lower threshold without exceeding the higher threshold, we use random sampling to try to find a concrete example $x\in P$  that violates the fairness property. If such a counterexample is found, we classify $P$ as \textit{undecided} and avoid splitting it further.
%\end{itemize}
%
Otherwise, we keep splitting $P$ into smaller partitions.

\begin{comment}
Our method differs from the vast majority of existing techniques for neural network verification, which focus on robustness verification of a single neural network~\cite{wang_formal_2018, shiqi2018neurify, singh_abstract_2019}. In contrast, our method focuses on certifying individual fairness by simultaneously executing the network twice symbolically, once for $\forall x\in X$ and another for $\forall x'\in X$ (we further explain in Section~\ref{sec:motivation} why directly using existing robustness verifiers to certify fairness does not work in practice).
%
Our method also differs from prior works on differential verification~\cite{paulsen_reludiff_2020, paulsen2020neurodiff} or equivalence verification~\cite{eleftheriadis2022neural}, which focus on proving the behavioral difference or equivalence of two different networks, given the same input. In our case, we deal with only one network model $f$, with different inputs defined by our fairness property.
\end{comment}

We have evaluated  our method on a large number of deep neural networks trained using four widely-used datasets for fairness research: Bank~\cite{misc_bank_marketing_222} (for predicting marketing), German~\cite{misc_statlog_(german_credit_data)_144} (for predicting credit risk), Adult~\cite{misc_adult_2} (for predicting earning power), 
and Compas~\cite{compas2016propublica} (for predicting recidivism risk).
%, and Salary~\cite{weisberg_salary} (for predicting future salary). 
%
For comparison, we apply \Fairify{}~\cite{biswas_fairify_2023} since it represents the current state-of-the-art in certifying individual fairness; we also apply $\alpha$-$\beta$-CROWN since it is currently the best robustness verifier for deep neural networks. 
Our results show that $\alpha$-$\beta$-CROWN is not effective in certifying individual fairness. As for \Fairify{},  our method \FairQuant{} significantly outperforms \Fairify{} in terms of both accuracy and speed for all DNN benchmarks. 
In fact, \FairQuant{} often completes certification in seconds, whereas \Fairify{} often times out after 30 minutes and certifies nothing or only a tiny fraction of the entire input domain.

To summarize, this paper makes the following contributions:
\begin{itemize}
\item We propose the first scalable method for certifying and quantifying individual fairness of DNNs using symbolic interval based analysis techniques.
  \begin{enumerate}
  \item For forward analysis,  we propose techniques for more accurately deciding if the DNN is fair/unfair for all inputs in an input partition. 
  \item For refinement, we propose techniques for more effectively deciding how to split the input partition. 
  \item For quantification, we propose techniques for efficiently computing the percentages of inputs whose outputs can be certified and falsified.
  \end{enumerate}
\item We demonstrate the advantages of our method over the current state-of-the-art on a large number of DNNs trained using four popular fairness research datasets.
\end{itemize}

The remainder of this paper is organized as follows.  First, we motivate our work in Section~\ref{sec:motivation} using examples.  Then, we present the technical background in Section~\ref{sec:prelims}.  Next, we present the high-level procedure of our method in Section~\ref{sec:method}, followed by detailed algorithms of the abstraction, refinement, and quantification subroutines in Sections~\ref{sec:subroutine1}, \ref{sec:subroutine2} and \ref{sec:subroutine3}. We present the experimental results in Section~\ref{sec:experiment}, review the related work in Section~\ref{sec:related}, and finally give our conclusions in Section~\ref{sec:conclusion}.

\section{Motivation}
\label{sec:motivation}

In this section, we use an example to illustrate the limitations of existing methods. 

\subsection{The Motivating Example}

\begin{figure}
    \centering
    \resizebox{\linewidth}{!}{
        \begin{tikzpicture}
\tikzset{every node/.style={font=\footnotesize}}

% Input layer
\node[circle, draw=black, fill=none, minimum size=20pt] (i1) at (0, 2) {$i_1$};
\node[circle, draw=black, fill=lightred, minimum size=20pt] (i2) at (0, 0) {$i_2$};
\node[circle, draw=black, fill=none, minimum size=20pt] (i3) at (0, -2) {$i_3$};

% Hidden layer
\foreach \h in {2,1}
\node[circle, draw=black, fill=none, minimum size=20pt] (h\h) at (1.5, 3-2*\h) {$h_\h$};

% Output layer
\node[circle, draw=black, fill=none, minimum size=20pt] (o) at (3, 0) {$o$};

\node[left=0cm of i1] {$i_1 \in [x_1, x_1]$};
\node[left=0cm of i2] {$\boldsymbol{i_2 \in [0, 0]}$};
\node[left=0cm of i3] {$i_1 \in [x_3, x_3]$};

% Connect nodes with edges
\draw[->] (i1) -- (h1) node[pos=0.1, right] {2.0};
\draw[->] (i1) -- (h2) node[pos=0.2, left] {-0.2};
\draw[->] (i2) -- (h1) node[pos=0.2, above] {0.5};
\draw[->] (i2) -- (h2) node[pos=0.2, below] {0.7};
\draw[->] (i3) -- (h1) node[pos=0.2, left] {1.2};
\draw[->] (i3) -- (h2) node[pos=0.1, right] {0.4};

\draw[->] (h1) -- (o) node[near end, above] {0.2};
\draw[->] (h2) -- (o) node[near end, below] {-1.0};

\end{tikzpicture}
        \begin{tikzpicture}
\tikzset{every node/.style={font=\footnotesize}}

% Input layer
\node[circle, draw=black, fill=none, minimum size=20pt] (i1) at (0, 2) {$i_1$};
\node[circle, draw=black, fill=lightred, minimum size=20pt] (i2) at (0, 0) {$i_2$};
\node[circle, draw=black, fill=none, minimum size=20pt] (i3) at (0, -2) {$i_3$};

% Hidden layer
\foreach \h in {2,1}
\node[circle, draw=black, fill=none, minimum size=20pt] (h\h) at (1.5, 3-2*\h) {$h_\h$};

% Output layer
\node[circle, draw=black, fill=none, minimum size=20pt] (o) at (3, 0) {$o$};

\node[left=0cm of i1] {$i_1 \in [x_1, x_1]$};
\node[left=0cm of i2] {$\boldsymbol{i_2 \in [1, 1]}$};
\node[left=0cm of i3] {$i_1 \in [x_3, x_3]$};

% Connect nodes with edges
\draw[->] (i1) -- (h1) node[pos=0.1, right] {2.0};
\draw[->] (i1) -- (h2) node[pos=0.2, left] {-0.2};
\draw[->] (i2) -- (h1) node[pos=0.2, above] {0.5};
\draw[->] (i2) -- (h2) node[pos=0.2, below] {0.7};
\draw[->] (i3) -- (h1) node[pos=0.2, left] {1.2};
\draw[->] (i3) -- (h2) node[pos=0.1, right] {0.4};

\draw[->] (h1) -- (o) node[near end, above] {0.2};
\draw[->] (h2) -- (o) node[near end, below] {-1.0};

\end{tikzpicture}
    }
    \caption{Symbolic interval analysis of an example DNN for making hiring decisions: the left figure is for female applicants ($i_2\in[0,0]$), and the right figure is for male applicants (where $i_2\in[1,1]$). Except for the protected attribute $i_2$, the symbolic intervals of the other attributes are the same.}
    \label{fig:example.network}
\end{figure}
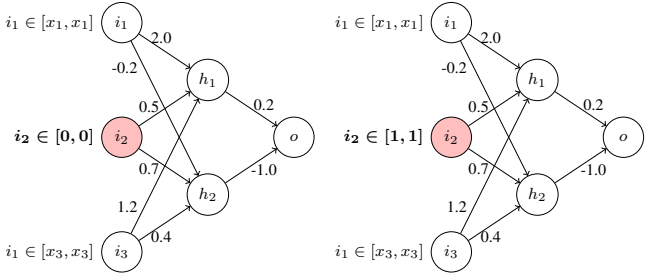

Fig.~\ref{fig:example.network} (left) shows a DNN for making hiring decisions.  It has three input nodes ($i_1,i_2$ and $i_3$), two hidden neurons ($h_1$ and $h_2$) and one output node ($o$). 
The values of $h_1$ and $h_2$ are computed in two steps:  first, the values of $i_1,i_2$ and $i_3$ are multiplied by the edge weights before they are added up; then, the result is fed to an activation function. For instance, the activation function may be ReLU$(z) = max(0,z)$. 
The output of the entire network $f$ is based on whether the value of $o$ is above 0; that is, positive label is generated if $o > 0$; otherwise, negative label is generated.

The DNN takes an input vector $x$ with three attributes: 
$x_1$ is the interview score of the job applicant, 
$x_2$ is the gender (0 for female and 1 for male), and 
$x_3$ is the number of years of experience. 
Furthermore, $x_2$ is the protected attribute while $x_1$ and $x_3$ are unprotected attributes.
In general, the input domain may be unbounded, e.g., when some attributes are real-valued variables.
However, for illustration purposes, we assume that the input domain is $X = \{x ~|~ x_1\in\{1,2,3,4,5\}, x_2\in\{0,1\}, \mbox{ and } x_3\in\{0,1,2,3,4,5\} \}$, meaning that $X$ has a total of $5\times 2\times 6 = 60$ individuals.

% We say that the DNN is \emph{individually fair} for a given input $x\in X$ if  we have $f(x) = f(x')$ (and unfair if $f(x) \neq f(x')$), where $x'$ is the result of flipping the protected attribute $x_2$ of the input $x$. 
% %
% The DNN is \emph{individually fair} for the entire input domain $X$ if we have $f(x) = f(x')$ (and unfair if $f(x) \neq f(x')$) for all $x\in X$.
% %
% % The DNN is \emph{unfair} for the input domain $X$ if we have $f(x) \neq f(x')$, for all $x\in X$.
% %
% Otherwise, the DNN remains \emph{undecided}.
% %
% This definition of fairness is motivated by the fact that we do not want a hiring algorithm to discriminate against people based on the protected attribute. 

Consider the individual $x = \langle 5, 0, 5\rangle$, meaning that $x_1=5$, $x_2=0$ and $x_3=5$.  According to the DNN in Fig.~\ref{fig:example.network}, the output is the positive label.  
After flipping the value of the protected attribute $x_2$ from 0 to 1, we have the individual $x' = \langle 5, 1, 5\rangle$, for which the DNN's output is also the positive label.
Since the DNN's output is oblivious to the gender attribute, we say that it is fair for this input $x$.

Consider another individual $x = \langle 1, 0, 5 \rangle$ whose gender-flipped counterpart is $x' = \langle 1, 1, 5 \rangle$. Since the DNN produces the negative label as output for both, it is still fair for this input $x$. 

To summarize, the DNN $f$ may be fair regardless of whether a particular $x \in X$ receives a positive or negative output; as long as $x$ receives the same label as its counterpart $x'$, the DNN is considered fair.

In contrast, since the individual $x = \langle 1, 0, 3 \rangle$ and its counterpart $x' = \langle 1, 1, 3 \rangle$ receive different outputs from $f$, where $x$ gets the positive label but $x'$ gets the negative label, the DNN is not fair for this input $x$.  Furthermore, this pair $(x, x')$ serves as a counterexample.

\begin{comment}
In general, a machine learning model may be fair for some, but not all, individuals of the input domain. 
%
% After all, we normally do not expect a machine learning model to be 100\% accurate, so we should not  expect the model to be fair for all inputs either.
%
However, this observation seems to be overlooked by prior work on certifying individual fairness, which focuses on searching for counterexamples or proving that they do not exist. 
%
Such qualitative techniques may not be useful if almost all of the DNNs in practice are simply declared as ``not fair''.
%
In contrast, we emphasize in this work the need to  \textit{quantitatively} measure the degree of fairness when a DNN is neither 100\% fair nor 100\% unfair, by computing the percentages of inputs whose classification outputs we can prove as fair or unfair. 
\end{comment}

\subsection{Limitations of Prior Work}

One possible solution to the fairness certification problem as defined above would be explicit enumeration of the $(x,x')$ pairs. For each $x\in X$, we may flip its protected attribute to generate $x'$ and then check if $f(x)=f(x')$.  
However, since the size of the input domain $X$ may be extremely large  or infinite, this method would be prohibitively expensive in practice. 

Another possible solution is to leverage existing DNN robustness verifiers, such as ReluVal~\cite{wang_formal_2018, shiqi2018neurify}, DeepPoly~\cite{singh_abstract_2019}, and $\alpha$-$\beta$-CROWN~\cite{wang2021beta}. However, since robustness and individual fairness are fundamentally different properties,  applying a robustness verifier would not work well in practice. 
The reason is because a robustness verifier takes an individual $x$ and tries to prove that small perturbation of $x$ (often defined by $||x-x'||<\delta$, where $\delta$ is a small constant) does not change the output label.
However, during fairness certification, we are not given a concrete individual $x$; instead, we are supposed to check for all $x \in X$ and $x' \in X$, where $x_j \neq x'_j$.
If we force a robustness verifier to take a symbolic input $I$ ($\forall x\in X$), it would try to prove that the DNN produces the same output label for all inputs in $X$ (implying that the DNN makes the same decision for all inputs in $X$).

Recall our example network $f$ in Fig.~\ref{fig:example.network}. While our method can prove that $f$ is fair for an input domain that contains $x=\langle 5,0,5 \rangle$ and $x'=\langle 5,1,5\rangle$ (both receive a positive outcome) as well as $x=\langle 1,0,5 \rangle$ and $x'=\langle 1,1,5\rangle$ (both receive a negative outcome), this cannot be accomplished by a robustness verifier (since it is almost never possible for all individuals in the input domain to have the same outcome).

The only currently available method for (qualitatively) certifying individual fairness of a DNN is \Fairify{}~\cite{biswas_fairify_2023}, which relies on the SMT solver and may return one of the following results:
SAT (meaning that there exists a counterexample that violates the fairness property),  UNSAT (meaning that there is no counterexample), or UNKNOWN (meaning that the result remains inconclusive). 
The main problem of \Fairify{} is that it works directly in the concrete domain by precisely encoding the non-linear computations inside the DNN as logical formulas and solving these formulas using the SMT solver.
Since each call to the SMT solver is NP-complete, the overall computational cost is high.
Although \Fairify{} attempts to reduce the computational cost by partitioning input domain \emph{a priori} and  heuristically pruning logical constraints, it does not scale as the network size and input dimension increase.  Indeed, our experimental evaluation of \Fairify{} shows that only tiny networks (with $\leq 100$ neurons) can be certified.
%, which is consistent with the results of the \Fairify{} paper.
%
%Another limitation is that \Fairify{} cannot perform \emph{quantitative} verification.

\subsection{Novelty of Our Method}

We overcome the aforementioned (accuracy and scalability) limitations by developing a method that is both \emph{scalable} and able to \emph{quantify} the degree of fairness.

First,  \FairQuant{} relies on abstraction to improve efficiency/scalability while maintaining the accuracy of symbolic forward analysis.  This increases the chance of quickly certifying more input regions as fair or falsifying them as unfair, and decreases the chance of leaving them as undecided.
Specifically, we use symbolic interval analysis, instead of the SMT solver used by \Fairify{}. The advantage is that symbolic interval analysis focuses on the behavior of the DNN in an abstract domain, which is inherently more efficient and scalable than analysis in the concrete domain.

Second, \FairQuant{} relies on iterative refinement (partitioning of the input domain) to improve the accuracy of forward analysis. Instead of creating input partitions \emph{a priori}, it conducts iterative refinement on a ``need-to'' basis guided by the fairness property to be certified.  
%That is, we partition an input region only when its certification result remains \emph{undecided}; at that moment, we select an input attribute $x_i$ and bisect its input interval $x_i\in(lb,ub)$, in order to maximize the chance of certifying or falsifying the individual partitions during the next iteration. 
%
This makes it more effective than the static partitioning technique of \Fairify{}, which divides the  input domain into a fixed number of equal chunks even before verifying any of them.

To see why iterative refinement can improve accuracy, consider our running example in Fig.~\ref{fig:example.network}.  Initially, forward analysis is applied to the DNN in the entire input domain $X$, for which the certification result is \emph{undecided}.
During refinement, our method would choose $x_1$ (over $x_3$) to split, based on its impact on the network's output.  After splitting $x_1\in\{1,2,3,4,5\}$ into $x_1\in\{1,2,3\}$ and $x_1\in\{4,5\}$, we apply forward analysis to each of these two new partitions.

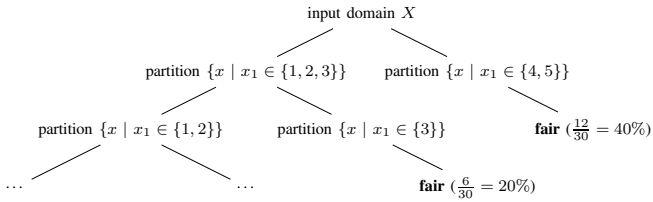
\begin{figure}
    \centering
    \resizebox{\linewidth}{!}{\begin{tikzpicture}[
    font=\footnotesize,
    level distance=1cm,
    level 1/.style={sibling distance=4cm},
    level 2/.style={sibling distance=2cm}
    level 3/.style={sibling distance=1cm}
]
% Define tree nodes
\node {input domain $X$}
    child {
        node {partition $\{x ~|~ x_1 \in \{1,2,3\} \}$}
        child {
            node {partition $\{x ~|~ x_1 \in \{1,2\}\}$}
            child {
                node {\ldots}
            }
            child {
                node {\ldots}
            }
        }
        child {
            node {partition $\{x ~|~ x_1 \in \{3\}\}$}
            child[edge from parent path={}] {
               node {}
            }
            child {
                node {\textbf{fair} ($\frac{6}{30} = 20\%$)}
            }
        }
    }
    child {
        node {partition $\{x ~|~ x_1 \in \{4,5\}\}$}
        child[edge from parent path={}] {
            node {}
        }
        child {
            node {\textbf{fair} ($\frac{12}{30} = 40\%$)}
        }
    };
\end{tikzpicture}}
    \caption{Iterative refinement tree for the example DNN in Fig.~\ref{fig:example.network}, to increase the chance of certifying or falsifying the DNN within an input partition.}
    \label{fig:tree}
\end{figure}

As shown in Fig.~\ref{fig:tree}, while the partition for $x_1\in\{1,2,3\}$ remains \emph{undecided}, the  partition for $x_1\in\{4,5\}$ is certified as \emph{fair}.   This partition has 12 pairs of $x \in X$ and $x' \in X$, where $x_2 \neq x'_2$. Therefore, from the input domain $X$ which has 30 $(x, x')$ pairs, we certify $12/30 = 40\%$ as \emph{fair}.
Next, we split the \emph{undecided} partition $x_1\in\{1,2,3\}$ into $x_1\in\{1,2\}$ and $x_1\in\{3\}$ and apply forward analysis to each of these two new partitions.  While the first new partition remains \emph{undecided}, the second one is certified as \emph{fair}.  Since this partition has six ($x, x'$) pairs, it represents $6/30=20\%$ of the input domain.

This iterative refinement process continues until one of the following two termination conditions is satisfied:  either there is no more partition to apply forward analysis to, or a predetermined time limit (e.g., 30 minutes) is reached.

\section{Preliminaries}
\label{sec:prelims}

In this section, we review the fairness definitions as well as the basics of neural network verification.

\subsection{Fairness Definitions}

Let $f:X\rightarrow Y$ be a classifier, where $X$ is the input domain and $Y$ is the output range.  Each input $x\in X$ is a vector in the $D$-dimensional attribute space, denoted $x=\langle x_1,\ldots,x_D\rangle$, where $1,\dots,D$ are vector indices.  Each output $y\in Y$ is a class label. 
Some attributes are legally protected attributes (e.g., gender and race) while others are unprotected attributes. Let $\mathcal{P}$ be the set of vector indices corresponding to protected attributes in $x$. We say that $x_j$ is a protected attribute if and only if $j \in \mathcal{P}$.

\begin{definition}[Individual Fairness for a Given Input]
Given a classifier $f$, an input $x \in X$, and a protected attribute $j \in \mathcal{P}$, we say that $f$ is individually fair for $x$ if and only if $f(x) = f(x')$ for any $x' \in X$ that differs from $x$ only in the protected attribute $x_j$. 
\end{definition}

This notion of fairness is local in the sense that it requires the classifier to treat an individual $x$ in a manner that is oblivious to its protected attribute $x_j$ of $x$. 
%
%Since it is local, by definition, fairness certification is as  simple as (1) flipping the protected attribute $x_j$ of $x$ to construct $x'$, (2) feeding $x$ and $x'$ to the classifier $f$ separately, and (3) checking if $f(x)=f(x')$.
%
% For example, if $x$ represents Alice, then $x'$ represents the gender-flipped counterpart, and $f(x)=f(x')$ means the decision made for Alice is individually fair with respect to gender.

\begin{definition}[Individual Fairness for the Input Domain]
Given a classifier $f$, an input domain $X$, and a protected attribute $j \in \mathcal{P}$, we say that $f$ is individually fair  for the input domain $X$ if and only if, for all $x\in X$,  $f(x)=f(x')$ holds for any $x'\in X$ that differs from $x$ only in the protected attribute $x_j$. 
\end{definition}

This notion of fairness is global since it requires the classifier to treat all $x\in X$ in a manner that is oblivious to the protected attribute $x_j$ of $x$. 
The method of explicit enumeration would be  prohibitively expensive since the number of individuals in $X$ may be astronomically large or infinite.

\subsection{Connecting Robustness to Fairness}

%At this moment, it is useful to compare individual fairness with 
Perturbation robustness, which is the most frequently checked property by existing DNN verifiers, is closely related to the notion of adversarial examples. The idea is that, if the DNN's classification output were robust, then applying a small perturbation to a given input $x$ should not change the classifier's output for $x$.

\begin{definition}[Robustness]
Given a classifier $f$, an input $x \in X$, and a small constant $\delta$,  we say that $f$ is robust against $\delta$-perturbation if and only if $f(x)=f(x')$ holds for all $x' \in X$ such that $||x-x'||\leq \delta$. 
\end{definition}

By definition, perturbation robustness is a local property defined for a particular input $x$, where the set of inputs defined by $||x-x'||\leq \delta$ is not supposed to be large. 
While in theory, a robustness verifier may be forced to check individual fairness by setting $\delta$ to a large value (e.g., to include the entire input domain $X$), it almost never works in practice. 
The reason is because, by definition, such a global robustness property requires that all inputs to have the same classification output returned by the DNN --  such a classifier $f$ would be practically useless.

This observation has been confirmed by our experiments using $\alpha$-$\beta$-CROWN~\cite{wang2021beta}, a state-of-the-art DNN robustness verifier.  Toward this end, we have created a merged network that contains two copies of the same network, with one input for one protected attribute group (e.g., male) and the other input for the other group (e.g., female). While the verifier finds counterexamples in seconds (and thus falsifies fairness of the DNN), it has the same limitation as \Fairify{}: it merely declares the DNN as \emph{unsafe} (unfair, in our context) as soon as it finds a counterexample, but does not provide users with any meaningful, quantitative information.  
In contrast, our method provides a quantitative framework for certified fairness by reasoning about all individuals in the input domain.

\section{Overview of Our Method}
\label{sec:method}

In this section, we present the top-level procedure of our method; detailed algorithms of the subroutines will be presented in subsequent sections.

Let the DNN $y = f(x)$ be implemented as a series of affine transformations followed by nonlinear activations, where each affine transformation step and its subsequent nonlinear activation step constitute a hidden layer. 
Let $l$ be the total number of hidden layers, then $f=f_l(f_{l-1}(...f_2(f_1(x \cdot W_1) \cdot W_2)... \cdot W_{l-1})$.  For each $k\in [1,l]$, $W_k$ denotes the affine transformation and $f_k()$ denotes the nonlinear activation.
More specifically, $W_1$ consists of the edge weights at layer 1 and $x \cdot W_1 = \Sigma_i x_iw_{1,i}$.  Furthermore, $f_1$ is the activation function, e.g.,  ReLU$(x \cdot W_1) = max(0, x \cdot W_1)$. 

\subsection{The Basic Components}

Similar to existing \emph{symbolic interval analysis} based DNN verifiers~\cite{wang_formal_2018,singh_abstract_2019,wang2021beta}, our method consists of three basic components: forward analysis, classification, and refinement.

\paragraph{Forward Analysis}

The goal of forward analysis is to compute upper and lower bounds of the network's output for all inputs. 
It starts by assigning a symbolic interval to each input attribute.  For example, in Fig.~\ref{fig:example.network}, $i_1=[x_1,x_1]$ is symbolic, where $x_1\in \{0,1,2,3,4,5\}$. 
%
%Interval bounds propagation is a type of abstract interpretation of the nonlinear arithmetic computations in the DNN.  
Compared to concrete values, symbolic values have the advantages of making the analysis faster and more scalable.  They are also sound in that they overapproximate the possible concrete values. 
%Thus, interval bounds propagation is well-suited for verification. 

\paragraph{Classification}

In a binary classifier, e.g., the DNN in Fig.~\ref{fig:example.network} for making hiring decisions, the output is a singular node $o$ whose numerical values needs to be turned to either the positive or the negative class label based on a threshold value (say 0).
%
%Now, consider the result of the forward analysis, which includes an upper bound and a lower bound for the output node $o$. 
%
For example, if $o \in [0.3, 0.4]$, the output label is guaranteed to be positive since $o>0$ always holds, 
and if $o \in [-0.7, -0.6]$, the output label is guaranteed to be negative since $o<0$ always holds.
%
%More formally, given our output interval $o \in [o_{lb}, o_{ub}]$, the output label is positive if $o_{lb} > 0$ and is negative if $o_{ub} < 0$.
%
However, if $o \in [-0.7, 0.4]$, the output label  remains \emph{undecided} -- this is when our method needs to conduct refinement.

\paragraph{Refinement}

The goal of refinement is to partition the input domain of the DNN to improve the (upper and lower) symbolic bounds computed by forward analysis. 
Since approximation error may be introduced when linear bounds are pushed through nonlinear activation functions (e.g., unless ReLU is always on or always off), by partitioning the input domain, we hope to increase the chance that activation functions behave similar to their linear approximations for each of the new (and smaller) input partitions, thus reducing the approximation error.

%In this context, the key is to identify which input attribute is making the most significant impact. In existing tools~\cite{wang_formal_2018}, this is often computed by analyzing the gradients of the network's output with respect to the network's input attributes.  However, such a technique is designed specifically for robustness verification. In our work, the context is different in that we have two separate forward analysis pass, one for $I$ ($\forall x\in X$) and the other for $I'$ ($\forall x'\in X$), as shown in Fig.~\ref{fig:example.network}.  
%
%Thus, we propose a new refinement strategy accordingly.

\subsection{The Top-Level Procedure}

Algorithm~\ref{alg:overall} shows the top-level procedure, whose input consists of the network $f$, the protected attribute $x_j$, and the input domain $X$. 
Together, these three parameters define the fairness certification problem, denoted $\langle f, x_j, X \rangle$.

\begin{algorithm}
{\footnotesize
\DontPrintSemicolon
\SetKwInput{KwData}{Input}
\caption{Overview Method of \FairQuant{}}
\label{alg:overall}
\KwData{neural network $f$, protected attribute $x_j$, input domain $X$}
\KwResult{$r_{cer}, r_{fal}, r_{und}$, which are percentages of certified, falsified, and undecided inputs}

Initial partition $P \leftarrow X$ with refinement depth 0

Push $P$ into an empty stack $S$

\tcp{initially, 100\% undecided}
$r_{cer} \leftarrow 0, r_{fal} \leftarrow 0, r_{und} \leftarrow 1$ 

%$cex\_count \leftarrow 0$

\While{$S$ is not empty and not-yet-timed-out}{

    Pop $P$ from the stack $S$    

    \tcp{certify current partition}
    $result  \leftarrow$ \textsc{SymbolicForward}($f, x_j, P$)  
    
    \If{result = Undecided}{
        \tcp{split current partition}
        $P_l, P_u$ $\leftarrow$ \textsc{BackwardRefinement}($f, P$) 

        Push $P_l$ and $P_u$ into the stack $S$
    }

    \tcp{update the percentages}
    $r_{cer}, r_{fal}, r_{und} \leftarrow$ \textsc{QuantifyFairness}($result, P, X$)  

}

%Increase $r_{fal}$ and decrease $r_{und}$ based on \textit{cex\_count} \tcp*{total counterexamples}

\Return{$r_{cer}, r_{fal}, r_{und}$}
}
\end{algorithm}

Within the top-level procedure, we first initialize the input partition $P$ as $X$ and push it into the stack $S$. Each input partition is associated with a refinement depth.  Since $P$ is initially the entire input domain $X$, its refinement depth is set to 0.  Subsequently, the refinement depth increments every time $P$ is bisected to two smaller partitions.  In general, the refinement depth of $P\subseteq X$ is the number of times that $X$ is bisected to reach $P$.

In Lines 4-10 of Algorithm~\ref{alg:overall}, we go through each partition stored in the stack $S$, until there is no partition left or a time limit is reached. For each partition $P$, we first apply symbolic forward analysis \textcolor{black}{(Line 6)} to check if the DNN $f$ is fair for all individuals in $P$. There are three possible outcomes:
(1) \emph{fair} (certified), meaning that $f(x)=f(x')$ for all $x\in P$ and its counterpart $x'$;
(2) \emph{unfair} (falsified), meaning that $f(x)\neq f(x')$ for all $x\in P$ and its counterpart $x'$; or
(3) \emph{undecided}.

Next, if the result is undecided (Line 7), we apply backward refinement by splitting $P$ into two disjoint new partitions $P_l$ and $P_u$.  By focusing on each of these smaller partitions in a subsequent iteration step, we hope to increase the chance of certifying it as fair (or falsifying it as unfair).

Finally, we quantify fairness \textcolor{black}{(Line 10)} by updating the percentages of certified ($r_{cer}$), falsified ($r_{fal}$) and undecided ($r_{und}$) inputs of $X$.
Specifically, if the previously-undecided partition $P$ is now certified as \emph{fair}, we decrease the undecided rate $r_{und}$ by $|P|/|X|$ and increase the certified rate $r_{cer}$ by the same amount.
On the other hand, if $P$ is falsified as \emph{unfair}, we decrease $r_{und}$ by $|P|/|X|$ and increase the falsified rate $r_{fal}$ by the same amount.

% The entire while-loop terminates when all partitions in the stack $S$ are decided, or a predetermined time limit is reached.

In the next three sections, we will present our detailed algorithms for forward analysis (Section~\ref{sec:subroutine1}), backward refinement (Section~\ref{sec:subroutine2}), and quantification (Section~\ref{sec:subroutine3}).

\subsection{The Correctness}

Before presenting the detailed algorithms, we would like to make two claims about the correctness of our method.
The first claim is about the \emph{qualitative} result of forward analysis, which may be fair, unfair, or undecided.

\begin{theorem}
When forward analysis declares an input partition $P\subseteq X$ as \emph{fair}, the result is guaranteed to be sound in that $f(x)=f'(x)$ holds for all $x\in P$ and its counterpart $x'$, 
Similarly, when forward analysis declares $P$ as \emph{unfair}, the result is guaranteed to be sound in that 
 $f(x)\neq f'(x)$ holds for all $x\in P$ and its counterpart $x'$.
\end{theorem}

The above soundness guarantee is because \textsc{SymbolicForward} soundly overapproximates the DNN's actual behavior.
That is, the upper bound $UB$ is possibly-bigger than the actual value, and the lower bound $LB$ is possibly-smaller than the actual value. As a result, the symbolic interval $[LB, UB]$ computed by \textsc{SymbolicForward} guarantees to include all concrete values.
In the next three sections, we shall discuss in more detail how the symbolic interval is used to decide if $P$ is \emph{fair}, \emph{unfair}, or \emph{undecided}.

When an input partition $P$ is \emph{undecided}, it means that some individuals in $P$ may be treated fairly whereas others in $P$ may be treated unfairly.
%
% Generally speaking, the percentage of \emph{fair} inputs reported by our method is a lower bound, and the percentage of \emph{unfair} inputs is also a lower bound.
%
This brings us to the second claim about the \emph{quantitative} result of our method, represented by the rates $r_{cer}$, $r_{fal}$ and $r_{und}$.

\begin{theorem}
The certification rate $r_{cer}$ computed by our method is guaranteed to be a lower bound of the percentage of inputs whose outputs are actually fair. 
Similarly, the falsified rate $r_{fal}$ is a lower bound of the percentage of inputs whose outputs are actually unfair.
\end{theorem}

In other words, when our method generates the percentages of \emph{fair} and \emph{unfair} inputs, it guarantees that they are provable lower bounds of certification and falsification, respectively.
The reason is because \textsc{SymbolicForward} soundly overapproximates the actual value range. When the output intervals indicate that the model is \emph{fair} (\emph{unfair}) for all inputs in $P$,  it is definitely \emph{fair} (\emph{unfair}). Thus, both $r_{cer}$ and $r_{fal}$ are guaranteed to be lower bounds.

Since the sum of the three rates is 1, meaning that $r_{und} = 1 - r_{cer} - r_{fal}$, the undecided rate $r_{und}$ is guaranteed to be an upper bound.

\section{Symbolic Forward Analysis}
\label{sec:subroutine1}

Algorithm~\ref{alg:forward} shows our forward analysis subroutine, which takes the subproblem $\langle f, x_j, P \rangle$ as input and returns the certification result as output.  

\begin{algorithm}
{\footnotesize
\DontPrintSemicolon
\SetKwInput{KwData}{Input}
\caption{Subroutine \textsc{SymbolicForward}()}
\label{alg:forward}
\KwData{neural network $f$, protected attribute $x_j$, input partition $P$}
\KwResult{certification $result$, which may be fair/unfair/undecided}

%$R[numLayer][layerSize]; R'[numLayer][layerSize];$

$I \leftarrow P |_{x_j\in[0,0]}$   and 
$I' \leftarrow P |_{x_j\in[1,1]}$

$O \leftarrow$ \textsc{ForwardPass}($f,I$)  
     \tcp*{for $x\in P$ with $x_j\in[0,0]$}

$O' \leftarrow$ \textsc{ForwardPass}($f,I'$)
     \tcp*{for $x'\in P$ with $x_j\in[1,1]$}

\If{ ($O_{lb} > 0  \wedge O'_{lb} >0)  \vee (O_{ub}<0 \wedge O'_{ub}<0)$ }{ 
         $result \leftarrow$ Fair
}
\ElseIf{ $(O_{lb} > 0  \wedge O'_{ub} <0)  \vee (O_{ub}<0 \wedge O'_{lb}>0)$ }{ 
         $result \leftarrow$ Unfair
}
\Else{
        $result \leftarrow$ Undecided
}

\Return $result$
}
\end{algorithm}

\subsection{The Two Steps}

Our forward analysis consists of two steps.  
First, a standard \emph{symbolic interval} based analysis is invoked twice, for the symbolic inputs $I$ and $I'$, to compute the corresponding symbolic outputs $O$ and $O'$.  
Second, $O$ and $O'$ are used to decide if the certification result is \emph{fair}, \emph{unfair}, or \emph{undecided}.

In the first step, the symbolic input $I = P|_{x_j\in[0,0]}$ is defined as the subset of input partition $P$ where all inputs have the protected attribute $x_j$ set to 0. 
In contrast, $I' = P|_{x_j\in[1,1]}$ is defined as the subset of $P$ where all inputs have $x_j$ set to 1. 
The output $O$ is a sound overapproximation of $f(x)$ for $x\in I$, whereas the output $O'$ is a sound overapproximation of $f(x')$ for $x'\in I'$.
The subroutine \textsc{ForwardPass} used to compute $O$ and $O'$ is similar to any state-of-the-art neural network verifier based on symbolic interval analysis; in our implementation, we used the algorithm of ReluVal~\cite{wang_formal_2018}.

In the second step,  the two output intervals,
 $O=[O_{lb},O_{ub}]$ and 
 $O'=[O'_{lb},O'_{ub}]$, 
are used to compute the certification result. 
To understand how it works, recall that in the concrete domain, the numerical value of the DNN's output node is compared against a threshold, say 0, to determine if the output label should be positive or negative.
In the symbolic interval abstract domain,  the upper and lower bounds of the numerical values are used to determine if the model is fair, unfair, or undecided.

\begin{figure}
\centering
    \resizebox{\linewidth}{!}{
\scalebox{0.5}{\begin{tikzpicture}
    \draw (0, 2) -- (0, -1);
    \node[label=above:0] at (0,2) {};

    % Both negative
    \draw (-2.5,1.5) -- (-2.5,1);
    \draw (-2.5,1) -- (-1.6,1) node[midway, below] {$O$};
    \draw (-1.6,1.5) -- (-1.6,1);

    \draw (-1.4,1.5) -- (-1.4,1);
    \draw (-1.4,1) -- (-0.5,1) node[midway, below] {$O'$};
    \draw (-0.5,1.5) -- (-0.5,1);

    \node[label=below:both negative] at (-1.5,0.5) {};
    \node[label=below:(fair)] at (-1.5,0.1) {};
    
    % Both positive
    \draw (2.5,1.5) -- (2.5,1);
    \draw (2.5,1.0) -- (1.6,1) node[midway, below] {$O'$};
    \draw (1.6,1.5) -- (1.6,1);

    \draw (1.4,1.5) -- (1.4,1);
    \draw (1.4,1) -- (0.5,1) node[midway, below] {$O$};
    \draw (0.5,1.5) -- (0.5,1);

    \node[label=below:both positive] at (1.5,0.5) {};
    \node[label=below:(fair)] at (1.5,0.1) {};
    
\end{tikzpicture}}
        \hspace*{0.05\linewidth}
\scalebox{0.5}{\begin{tikzpicture}
    \draw (0, 2) -- (0, -1);
    \node[label=above:0] at (0,2) {};

    % o' positive, o negative
    \draw (-1.4,1.5) -- (-1.4,1);
    \draw (-1.4,1) -- (-0.5,1) node[midway, below] {$O$};
    \draw (-0.5,1.5) -- (-0.5,1);

    \draw (1.4,1.5) -- (1.4,1);
    \draw (1.4,1) -- (0.5,1) node[midway, below] {$O'$};
    \draw (0.5,1.5) -- (0.5,1);

    \node[label=right:{$O$ neg, $O'$ pos}] at (1.5,1.5) {};
    \node[label=right:{(unfair)}] at (2,1.1) {};
    
    % o' negative, o positive
    \draw (-1.4,-0.5) -- (-1.4,0);
    \draw (-1.4,-0.5) -- (-0.5,-0.5) node[midway, below] {$O'$};
    \draw (-0.5,-0.5) -- (-0.5,0);

    \draw (1.4,-0.5) -- (1.4,0);
    \draw (1.4,-0.5) -- (0.5,-0.5) node[midway, below] {$O$};
    \draw (0.5,-0.5) -- (0.5,0);

    \node[label=right:{$O$ pos, $O'$ neg}] at (1.5,0) {};
    \node[label=right:{(unfair)}] at (2,-0.4) {};
\end{tikzpicture}}
    }
    \caption{Sufficient conditions for deciding fairness based on symbolic output intervals $O$ and $O'$, and the threshold 0: there are two fair conditions (left) and two unfair conditions (right).}
    \label{fig:fairness}
\end{figure}
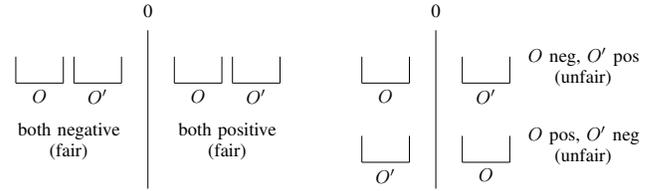

Below are the five scenarios:
\begin{enumerate}
    \item If $O_{lb} > 0$ and $O'_{lb} > 0$,  both $O$ and $O'$ have the positive label, meaning that $f$ is fair for $P$.
    \item If $O_{ub} < 0$ and $O'_{ub} < 0$,  both $O$ and $O'$ have the  negative label, meaning that $f$ is fair for $P$.
    \item If $O_{lb} > 0$ and $O'_{ub} < 0$,  $O$ is positive but $O'$ is negative, meaning that $f$ is unfair for $P$.
    \item If $O_{ub} < 0$ and $O'_{lb} > 0$,  $O$ is negative but $O'$ is positive, meaning that $f$ is unfair for $P$.
    \item Otherwise,  $f$ remains undecided for $P$.
\end{enumerate}
Fig.~\ref{fig:fairness} illustrates the first four scenarios above. Specifically, the horizontal line segments represent the value intervals of $O$ and $O'$, whose upper/lower bounds may be either $>0$ or $<0$.  The vertical lines represent the threshold value 0. 

We can extend out method from two protected attribute (PA) groups (e.g., male and female) to more than two PA groups.
For example, if the protected attribute $x_j$ has three values, we will have three symbolic inputs ($I,I'$ and $I''$) and three corresponding symbolic outputs ($O,O'$ and $O''$).
To decide if $f$ is fair (or unfair) in this multi-PA group setting, we check if (1) individuals in each PA group receive the same output label; and (2) the output labels for the three PA groups are the same.

We can also extend our method from binary classification to multi-valued classification.
For example, if there are three possible output labels, we will have $O_1, O_2,$ and $O_3$ as the symbolic intervals for the three values for one PA group ($I$) and $O'_1, O'_2$ and $O'_3$ for the other PA group ($I'$).
To decide if $f$ is fair (or unfair) for this multi-valued classification, we check (1) which output labels are generated for  $I$ and $I'$; and (2) whether these two output labels ($O$ and $O'$) are the same.

\subsection{The Running Example}

For our running example in Fig.~\ref{fig:example.network}, consider the initial input partition $P = X$. For ease of understanding, we denote the symbolic expressions for a neuron $n$ as $S_{in}(n)$ after the affine transformation, and as $S(n)$ after the ReLU activation. 
Furthermore, $S$ will be used for $I$, and $S'$ will be used  for $I'$.

Let 
$I = P|_{x_j=0}$ and $I' = P|_{x_j=1}$.
After affine transformation in the hidden layer, we have $S_{in}(h_1) = 2x_1 + 1.2x_3$ and $S'_{in}(h_1) = 2x_1 + 1.2x_3 + 0.5$. If we concretize these symbolic expressions, we will have $S_{in}(h_1) = [2, 16]$ and $S'_{in}(h_1) = [2.5, 16.5]$. Based on these concrete intervals, we know that $h_1$ is always active for both $I$ and $I'$. Since the activation function is ReLU, we have $S(h_1) = S_{in}(h_1)$ and $S'(h_1) = S'_{in}(h_1)$.

For the hidden neuron $h_2$, we have $S_{in}(h_2) = -0.2x_1 + 0.4x_3$ and $S'_{in}(h_2) = -0.2x_1 + 0.4x_3 + 0.7$, whose corresponding concrete bounds are $[-1, 1.8]$ and $[-0.3, 2.5]$, respectively. 
In both cases, since $h_2$ is nonlinear (neither always-on nor always-off), we must approximate the  values using linear expressions to obtain $S(h_2)$ and $S'(h_2)$. 
While we  use the sound overapproximation method of Wang et al.~\cite{shiqi2018neurify},  other techniques (e.g., \cite{ehlers2017formal, singh_abstract_2019}) may also be used. 

After overapproximating the ReLU behavior of $h_2$, we obtain $S(h_2) = [-0.128 x_1 + 0.257 x_3$, $-0.128 x_1 + 0.257 x_3 + 0.643]$ and $S'(h_2) = [-0.178 x_1 + 0.357 x_3 + 0.625$, $-0.178 x_1 + 0.357 x_3 + 0.893]$.

Finally, we compute  $S_{in}(o) = [0.528 x_1 - 0.017 x_3 - 0.643$, $0.528 x_1 - 0.017 x_3]$ and $S'_{in}(o) = [0.578 x_1 - 0.117 x_3 - 0.793$, $0.578 x_1 - 0.117 x_3 - 0.525]$. From these symbolic bounds, we  obtain the concrete bounds of $O = [-0.2, 2.64]$ and $O' = [-0.8, 2.368]$. Since these output intervals are not tight enough to determine the output labels for $I$ and $I'$, which are needed to decide if the model is fair or unfair for the partition $P$, the model remains undecided. 

To improve the accuracy, we need to split $P$ into smaller input partitions and then apply symbolic forward analysis to each partition again. How to split $P$ will be addressed by the iterative backward refinement method presented in the next section.

\section{Iterative Backward Refinement}
\label{sec:subroutine2}

The goal of iterative backward refinement is to split the currently-undecided input partition $P$ into smaller partitions, so that for each of these smaller partitions, symbolic forward analysis will obtain a more accurate result. Algorithm~\ref{alg:backward} shows the pseudo code, which takes the network $f$ and the partition $P$ as input and returns two smaller partitions $P_l$ and $P_u$ as output. Inside this procedure, Lines~7-14 are related to splitting $P$, and Lines 2-6 are related to early termination conditions.
%
%Specifically, if the refinement depth $P.depth$ exceeds some predefined thresholds, meaning that we have split too many times, we avoid splitting $P$ further to ensure quick termination.

\begin{algorithm}
{
\footnotesize
\DontPrintSemicolon
\SetKwInput{KwData}{Input}

% \captionsetup{font=footnotesize}
\caption{Subroutine \textsc{BackwardRefinement}()}

\label{alg:backward}
\KwData{neural network $f$, input partition $P$}
\KwResult{smaller partitions $P_l$ and $P_u$, if any}

%Let $R$ and $R'$ be gradient masks computed by \textsc{ForwardPass} for $I$ and $I'$

Let \textit{cex\_count} be the total number of counterexamples found

\If{($P.depth \geq$ \textit{max\_refinement\_depth})}{
    \Return  (null,null)  \tcp*{do not  split $P$}
}
\ElseIf{($P.depth \geq$ \textit{min\_sample\_depth}) and \textsc{SampledCEX}($P$)}{
    \textit{cex\_count} += 1 \\
    \Return  (null,null)   \tcp*{do not split $P$}
} 
\Else {

    \tcp{$R$, $R'$ are gradient masks computed for $I$, $I'$}
    $g_I \leftarrow$ \textsc{BackwardPass}($f,R$)
    
    $g_{I'} \leftarrow$ \textsc{BackwardPass}($f,R'$)
    
    $g  = ( g_I  + g_{I'} )/2$
    
    \tcp{Best input attribute to bisect}
    $x_i \leftarrow \mathbf{argmax}_{x_i} ~ g(x_i)*|ub(x_i)-lb(x_i)|$ 
    
    $P_l \leftarrow  P ~|~_{ x_i\in [lb, (lb+ub)/2] }$ 
    
    $P_u \leftarrow  P ~|~_{ x_i\in [(lb+ub)/2, ub] }$
    
    \Return $P_l, P_u$
}
}
\end{algorithm}

%\vspace{-1em}

\subsection{Early Termination Conditions}

In Lines 2-6 of Algorithm~\ref{alg:backward}, we  check if $P.depth$ exceeds the predefined \textit{max\_refinement\_depth}. If the answer is yes,  we avoid splitting  $P$ further. For example, if \textit{max\_refinement\_depth}$=20$, it means the current partition $P$ occupies only $|P|/|X|= \frac{1}{2^{20}}$ of the entire input domain $X$. 
By increasing the refinement depth, we can decrease the percentage of undecided inputs over $X$.

If $P.depth$ has not exceeded the maximal refinement depth, we check if $P.depth$ exceeds the predefined \textit{min\_sample\_depth}, which is set to a value (e.g., 15) smaller than \textit{max\_refinement\_depth}.  When $P.depth$ exceeds this threshold, we start searching for counterexamples in $P$ via random sampling. 

Inside the random sampling subroutine \textsc{SampledCEX}($P$) shown in Line~4, we sample up to 10 concrete inputs in $P$ and check if $x$ and its counterpart $x'$ satisfy $f(x) \neq f(x')$. If this condition is satisfied, a counterexample is found (but $P$ remains undecided); in this case, we increment \textit{cex\_count} and stop splitting $P$. If no counterexample is found, we continue splitting $P$ into smaller partitions.

Note that in both early termination cases (Lines 3 and 6), the partition $P$ will be marked as \textit{undecided} since  we are not able to decide whether the DNN model is fair or unfair to \emph{all individuals} in $P$.

\subsection{Splitting Input Intervals}

In Lines 7-14 of Algorithm~\ref{alg:backward}, we split $P$ into smaller partitions $P_l$ and $P_u$ by first identifying the input attribute $x_i$ that has the largest influence on the output (Lines 8-12) and then bisecting its input interval $x_i\in[lb,ub]$.

Our method for identifying the input attribute $x_i$ is based on maximizing the impact of an input attribute on the network's output.  One way to estimate the impact is taking the product of the gradient $g(x_i)$ and the input range $|ub(x_i)-lb(x_i)|$.  In the literature, the product is often called the \emph{smear value}~\cite{Kearfott:1990:AIP, Kearfott:1996:RGS}.
Unlike existing methods such as Wang et al.~\cite{wang_formal_2018}, however, our computation of the smear value is different because we must consider both inputs $I$ and $I'$, which may have different gradients.

Specifically, during forward analysis, we store the neuron activation information in two gradient mask matrices denoted $R$ and $R'$, where $R[i][j]$ is [1,1] if the $j$-th neuron at $i$-th layer is always active, [0,0] if it is always inactive, and [0,1] if it is unknown.  The neuron activation information is used later to perform backward refinement for this partition $P$.

During refinement, we first compute the two gradients $g_I$ and $g_{I'}$ and then take the average. Our goal is to identify the input attribute that has the largest \emph{overall} influence on the network's output.

%\input{algorithms/alg_gradient.tex}

%We initialize the gradients $g$ and $g'$ to the edge weights in the output layer. Then, for every layer moving backwards, we first perform ReLU backwards for every neuron based on its behavior indicated by $R$ (and $R'$) and then multiply by the weights of its incoming edges. When we obtain $g$ and $g'$ at the input layer, we take the average to obtain $g$. 
%
%This gradient, however, is not sufficient on its own to indicate the degree of influence. In fact, it is contingent upon the range of the input attribute. Therefore, we choose the feature with the largest smear value~\cite{Kearfott:1990:AIP, Kearfott:1996:RGS}, which is the range multiplied by the gradient at a given input feature. The resulting two smaller partitions $P_l$ and $P_u$ are then added to our stack $S$.

\subsection{The Running Example}

Consider our running example in Fig.~\ref{fig:example.network} again.  
%$h_1$ is always active and $h_2$ is nonlinear for both forward passes for $I$ and $I'$. With this information, we can conduct a backward analysis, starting from the second to last layer. 
To compute the smear value, we start with the output layer's edge weights, which are 0.2 for $h_1$ and -1 for $h_2$.
Since the ReLU associated with $h_1$ is always-on, $g_I(h_1)$ and $g_{I'}(h_1)$ are set to the interval $[0.2,0.2]$.
However, since the ReLU associated with $h_2$ is nonlinear, as indicated by the gradient mask matrices $R$ and $R'$, $g_I(h_2)$ and $g_{I'}(h_2)$ are set to the interval $[-1.0, 0]$.

Then, we propagate these gradient intervals backwardly, to get $g_I(i_1) = g_{I'}(i_1) = [(0.2\times2)+(0\times-0.2), (0.2\times2)+(-1\times-0.2)] = [0.4, 0.6]$ and $g_I(i_2) = g_{I'}(i_2) = [-0.6, 0.1]$ and $g_I(i_3) = g_{I'}(i_3) = [-0.16, 0.24]$. 

Next, we compute the average $g$, based on which we compute the smear values. Since $x_1$ has the smear value of $0.6 \times 4 = 2.4$  and $x_3$ has the smear value of $0.24 \times 5 = 1.2$, we choose to partition $P$ by bisecting the input interval of $x_1$. 
This leads to the smaller partitions shown in  Fig.~\ref{fig:tree}.

\subsection{Generalization}

While we only consider ReLU networks in this paper, our refinement technique can be extended to non-ReLU activations. Recall that, by definition, ReLU$(z)=0$ (inactive) if $z<0$, and ReLU$(z)=1$ (active) if $z>0$. Let $\sigma(z)$ be a non-ReLU activation function. To compute the gradient mask matrices $R$ and $R'$, we use thresholds ($\epsilon_1,\epsilon_2$) to approximate the on/off behavior:
the mask is [0,0] (inactive) if $z<\epsilon_1$ and [1,1] (active) if $z>\epsilon_2$.

Although the approximate on/off behavior of non-ReLU activation function $\sigma(z)$ is not the same as the on/off behavior of ReLU$(z)$, it serves as a practically-useful heuristic to rank the input attributes. Furthermore, this generalization will not affect the soundness of our method, since the gradient masks computed in this manner are only used for picking which input attribute to split first.

\section{Fairness Quantification}
\label{sec:subroutine3}

We now present our method for updating the percentages of certified and falsified inputs, when the DNN model is found to be fair or unfair for the current input partition $P$.  The pseudo code is shown in Algorithm~\ref{alg:rates}.

\begin{algorithm}
{\footnotesize
\DontPrintSemicolon
\SetKwInput{KwData}{Input}
\caption{Subroutine \textsc{QuantifyFairness}()}
\label{alg:rates}
\KwData{certification $result$, input partition $P$, input domain $X$}
\KwResult{Percentages $r_{cer}, r_{fal}, r_{und}$ }

\textit{partition\_size} $= \prod_{\forall x_i \in P, x_i \neq x_j} (UB(x_i) - LB(x_i))$

\textit{domain\_size} $= \prod_{\forall x_i \in X, x_i \neq x_j} (UB(x_i) - LB(x_i))$

\tcp{percentage of inputs in partition $P$}
$r_P$ = \textit{(partition\_size / domain\_size)}   

\If{$result =$ Fair}{
    $r_{cer} \mathrel{+}= r_p$ 

    $r_{und} \mathrel{-}= r_p$
}
\ElseIf{$result =$ Unfair}{
    $r_{fal} \mathrel{+}= r_P$ 

    $r_{und} \mathrel{-}= r_P$
}
\Return{$r_{cer}, r_{fal}, r_{und}$}
}
\end{algorithm}

There are three cases.  
First, if the current partition $P$ is found to be fair, meaning that all inputs in $P$ are treated fairly,  we compute the percentage of input domain $X$ covered by the partition $P$, denoted $r_{P}$, and then add $r_P$ to $r_{cer}$, the percentage of certified inputs.
Second, if the current partition $P$ is found to be unfair, meaning that all inputs in $P$ are treated unfairly, we add $r_P$ to $r_{fal}$, the percentage of falsified inputs.
In both cases, we also subtract $r_p$ from $r_{und}$. 
Otherwise, the current partition $P$ remains undecided and the percentages remain unchanged. %this percentage to our certification or falsification rate, for fair or unfair respectively, and subtract it from our undecided rate.
%

% The percentage $r_p$ is defined as the size of partition $P$ divided by the size of the input domain $X$. In general, both $X$ and $P$ may have an extremely large (and sometimes infinite) number of inputs.

Consider our running example with input partition $P$ defined as $x_1 \in [4,5] \wedge x_2 \in [0,1] \wedge  x_3 \in [0,5]$,  as shown by the right child of the root node in Fig.~\ref{fig:tree}. This partition has a total of 24 individuals, and its corresponding $I = P|_{x_2 = 0}$ and $I'= P|_{x_2 = 1}$ contain 12 individuals each. In contrast, the entire input domain $X$ has 60 individuals, or 30 pairs of $x$ and its counterpart $x'$ (where $x_2 \neq x'_2$). 

For this input partition $P$,  $O = [1.49, 2.65]$ and $O' = [1.0, 2.36]$ are the output intervals.   Assuming that the decision threshold is 0, the bounds of $O$ and $O'$ imply that the DNN model will generate the positive label for both $I$ and $I'$, meaning that the DNN model is fair for all individuals in $P$.

Since the input partition size is 12 and the input domain size is 30, the rate $r_{P} = \frac{12}{30} = 40\%$.  After certifying $P$ to be fair, we can add 40\% to $r_{cer}$, the certification rate, and consequently subtract 40\% from $r_{und}$, the undecided rate.

While the above computation assumes that population distribution for each feature is uniform and thus the percentage (e.g., 40\%) is computed directly from the partition size (e.g., 12) and the domain size (e.g., 30), the method can be easily extended to consider a non-uniform population distribution.
Furthermore, note that the method works regardless of whether the input attributes have integer or real values.

\section{Experiments}
\label{sec:experiment}

We have implemented \FairQuant{} in a software tool written in C, by leveraging the OpenBlas\footnote{http://www.openblas.net} library for fast matrix multiplication and symbolic representation of the upper and lower bounds.  Our forward analysis follows that of Wang et al.~\cite{wang_formal_2018, shiqi2018neurify}.
For experimental comparison, we also run \Fairify{} which is the only currently-available tool for DNN individual fairness certification. 
Since \Fairify{} cannot quantify the degree of fairness, we compute the certified/falsified/undecided rates based on its reported statistics.

It is worth noting that \Fairify{} and our method (\FairQuant{}) have a fundamental difference in falsification. \Fairify{} stops and declares an input partition as SAT as soon as it finds a  counterexample in that partition; thus, the number of counterexamples that it finds is always the same as the number of SAT partitions it reports. However, SAT partitions are not necessarily \emph{unfair} partitions, since \emph{unfair} partitions require all inputs to be counterexamples, but a SAT partition, excluding the one counterexample, still remains \textit{undecided}.

\FairQuant{} checks if an entire partition is \textit{unfair}. Moreover, when the partition is \textit{undecided}, it can minimize the amount of \textit{undecided} inputs by only sampling for counterexamples after it reaches a deep enough refinement depth. This is made possible through our iterative refinement.

For example, in a DNN model named GC-3, \Fairify{} finds 194 SAT partitions (together with 6 UNSAT and 1 UNKNOWN partitions).  However, none of these 194 SAT partitions are \emph{unfair} partitions. 
Instead, the percentage of falsified inputs is close to being 0\% (representing 194 counterexamples out of over 435 trillion individuals in the input domain), the percentage of certified inputs is 2.985\% (6 UNSAT partitions out of 201 partitions), and the rest remains undecided.
\FairQuant{}, on the other hand, finds 25,963 counterexamples; furthermore, it is able to formally certify 58.44\% of the inputs as \emph{fair}.

\begin{table}

\centering
\caption{Statistics of the datasets and DNNs used in our experiments.}
\label{table.model}

\resizebox{\linewidth}{!}{ 
\renewcommand{\arraystretch}{1.2}
\scriptsize

\begin{tabular}{| c|c|| c|c|c|c|c|c |} 

\hline
\textbf{Dataset (PA)} & \textbf{\# Inputs} & \textbf{DNN} & 
\textbf{\# Layers} & \textbf{\# Neurons} & \textbf{Accuracy (\%)} \\

\hline
\hline
\multirow{8}{*}{Bank (age)}
& \multirow{8}{*}{16}
& BM-1~\cite{biswas_fairify_2023} & 2 & 80 & 89.20 \\ 
& & BM-2~\cite{biswas_fairify_2023} & 2 & 48 & 88.76 \\ 
& & BM-3~\cite{biswas_fairify_2023} & 1 & 100 & 88.22 \\ 
& & BM-4~\cite{biswas_fairify_2023} & 3 & 300 & 89.55 \\ 
& & BM-5~\cite{biswas_fairify_2023} & 2 & 32 & 88.90 \\
& & BM-6~\cite{biswas_fairify_2023} & 2 & 18 & 88.94 \\ 
& & BM-7~\cite{biswas_fairify_2023} & 2 & 128 & 88.70 \\ 
& & BM-8~\cite{biswas_fairify_2023} & 5 & 124 & 89.20 \\ 

\hline
\multirow{5}{*}{German (age)}
& \multirow{5}{*}{20}
& GC-1~\cite{biswas_fairify_2023} & 1 & 50 & 72.67 \\ 
& & GC-2~\cite{biswas_fairify_2023} & 1 & 100 & 74.67 \\ 
& & GC-3~\cite{biswas_fairify_2023} & 1 & 9 & 75.33 \\ 
& & GC-4~\cite{biswas_fairify_2023} & 2 & 10 & 70.67 \\ 
& & GC-5~\cite{biswas_fairify_2023} & 5 & 124 & 69.33 \\

\hline
\multirow{12}{*}{Adult (gender)}
& \multirow{12}{*}{13}
& AC-1~\cite{biswas_fairify_2023} & 2 & 24 & 85.24 \\ 
& & AC-2~\cite{biswas_fairify_2023} & 1 & 100 & 84.70 \\ 
& & AC-3~\cite{biswas_fairify_2023} & 1 & 50 & 84.52 \\ 
& & AC-4~\cite{biswas_fairify_2023} & 2 & 200 & 84.86 \\ 
& & AC-5~\cite{biswas_fairify_2023} & 2 & 128 & 85.19 \\
& & AC-6~\cite{biswas_fairify_2023} & 2 & 24 & 84.77 \\
& & AC-7~\cite{biswas_fairify_2023} & 5 & 124 & 84.85 \\
& & AC-8~\cite{biswas_fairify_2023} & 2 & 10 & 82.15 \\
& & AC-9~\cite{biswas_fairify_2023} & 4 & 12 & 81.22 \\
& & AC-10~\cite{biswas_fairify_2023} & 4 & 20 & 78.56 \\
& & AC-11~\cite{biswas_fairify_2023} & 4 & 40 & 79.25 \\
& & AC-12~\cite{biswas_fairify_2023} & 9 & 45 & 81.46 \\

\hline
\multirow{7}{*}{Compas (race)}
& \multirow{7}{*}{6}
& compas-1 & 2 & 24 & 73.46 \\ 
& & compas-2 & 5 & 124 & 72.82 \\ 
& & compas-3 & 3 & 600 & 72.98 \\ 
& & compas-4 & 9 & 90 & 72.98 \\ 
& & compas-5 & 10 & 2000 & 72.01 \\ 
& & compas-6 & 4 & 4000 & 73.95 \\ 
& & compas-7 & 10 & 10000 & 72.49 \\ 

% \hline
% \multirow{7}{*}{\rotatebox[origin=c]{0}{Salary (gender)}} 
% & salary-1 & 2 & 24 & 66.67\\ 
% & salary-2 & 5 & 124 & 83.33\\
% & salary-3 & 3 & 600 & 66.67\\
% & salary-4 & 9 & 90 & 66.67\\ 
% & salary-5 & 10 & 2000 & 66.67\\ 
% & salary-6 & 4 & 4000 & 66.67\\ 
% & salary-7 & 10 & 10000 & 66.67\\ 

\hline
\end{tabular}
\renewcommand{\arraystretch}{1.0}
}

\end{table}

\subsection{Benchmarks}

Table~\ref{table.model} shows the statistics of the benchmarks, including {32} deep neural networks trained on {four} popular datasets for fairness research.
Among the {32} networks, 25 came from \Fairify{}~\cite{biswas_fairify_2023} and the other {7} were trained by ourselves using TensorFlow. All of these networks have a single node in the output layer, to determine the binary classification result. 
{
Columns 1-2 show the name of each dataset with its considered protected attribute (PA) and the number of input attributes.
Columns 3-6 show the name of each DNN model, its number of hidden layers, number of hidden neurons, and classification accuracy.
The accuracy for DNNs trained on  \emph{Bank}, \emph{German}, and \emph{Adult} was provided by the \Fairify{} paper. For the models we trained using \emph{Compas}, we have reserved 10\% of the data for testing.
}
{
All the networks coming directly from \Fairify{} on \emph{Bank}, \emph{German}, and \emph{Adult} datasets are small, where the largest has only 200 hidden neurons. Moreover, most of them have only 1 or 2 hidden layers. Thus, we additionally trained much larger networks, with up to 10,000 hidden neurons, using the \emph{Compas} dataset. 
}

Details of the {four} datasets are given as follows.
\emph{Bank}~\cite{misc_bank_marketing_222} is a dataset for predicting if a bank client will subscribe to its marketing; it consists of 45,000 samples.
\emph{German}~\cite{misc_statlog_(german_credit_data)_144} is a dataset for predicting the credit risk of a person; it consists of 1,000 samples.
\emph{Adult}~\cite{misc_adult_2} is a dataset for predicting if a person earns more than \$50,000; it consists of 32,561 samples.
{Finally,}
\emph{Compas}~\cite{compas2016propublica} is a dataset for predicting the risk of recidivism; it consists of 6,172 samples.\footnote{We used the preprocessed \emph{Compas} data provided by~\cite{adebayo2016fairml}.} 
%
% Finally, \emph{Salary}~\cite{weisberg_salary} is a dataset for predicting if a person can earn more than \$25,000; it consists of 52 samples. 

We evaluate our method using three legally-protected input attributes.
For \emph{Bank} and \emph{German}, we use \textit{age}; 
{for \emph{Adult}}, we use \textit{gender}; 
and for \emph{Compas}, we use \textit{race}.\footnote{
For Bank and German, we use binarized \textit{age} attribute provided by~\Fairify{}. For Compas, we binarize \textit{race} attribute into \{white, non-white\} as done in~\cite{le2022survey}.
}
These are consistent with \Fairify{} and other prior works in the fairness research.

\subsection{Experimental Setup}

We ran all experiments on a computer with 2 CPU, 4GB memory, and Ubuntu 20.04 Linux operating system.   
We set a time limit of 30 minutes for each DNN model. 
Our experiments were designed to answer three research questions:
\begin{enumerate}
 \item 
Is \FairQuant{} more accurate than the current state-of-the-art in certifying individual fairness of a DNN model?
\item
Is \FairQuant{} more scalable than the current state-of-the-art in handling DNN models, especially when the network size increases? 
\item 
Is \FairQuant{} more effective than the current state-of-the-art in providing feedback, e.g., by quantitatively measuring the percentages of certified, falsified, and undecided inputs?
\end{enumerate}

\Fairify{} requires a parameter \textit{MS} (maximum size of an input attribute) based on which it creates a fixed number of input partitions prior to certification.
On the DNN models trained for \emph{Bank}, \emph{German}, and \emph{Adult}, we used the default \textit{MS} values (100, 100, and 10) for \Fairify{} to create 510, 201, and 16000 partitions, respectively.
On the new DNN models trained {for \emph{Compas}}, we set \textit{MS} to a small value of 2 to create {20 partitions} for \Fairify{}.
% 20 and 468 partitions, respectively.
%
This was done to maximize \Fairify{}'s performance such that it does not “choke” in verifying each input partition.

By default, \Fairify{} uses 100 seconds as  “soft timeout” for each input partition and uses 30 minutes as “hard timeout” for the entire DNN. This means that it takes at most 100 seconds to verify a single input partition, and if unsolved, it just moves to the next partition, until the entire 30 minutes runs out.

To run \FairQuant{}, we set the parameters \textit{min\_check\_depth} to 15 and \textit{max\_refinement\_depth} to 20 for all DNN models. We also use 30 minutes as ``hard timeout'', but \FairQuant{} always finished before the limit.

\begin{table}
\centering
\caption{Results for fairness certification: \Fairify{} vis-à-vis \FairQuant{}
%, where T/O means time out, M/O means memory out,  \textit{Cex?} shows if a \textit{Counterexample} is found, \textit{\#Cex} shows the number of counterexamples found, \textit{Cer\%} shows the \textit{Certified} percentage, \textit{Fal\%} shows the \textit{Falsified} percentage, and \textit{Und\%} shwos the \textit{Undecided} percentage.
}
\label{table.results}

\resizebox{\linewidth}{!}{  
\renewcommand{\arraystretch}{1.2}
\Large

\begin{tabular}{ | cc || rcr|rrr || rcr|rrr | } 

\hline

\multirow{2}{*}{Dataset} & \multirow{2}{*}{DNN} & \multicolumn{6}{c||}{\textbf{Fairify}~\cite{biswas_fairify_2023}} & \multicolumn{6}{c|}{\textbf{FairQuant} (new)} \\

\cline{3-14}

&
& Time & Cex & \#Cex          & Cer\% & Fal\% & Und\% 
& Time & Cex & \#Cex          & Cer\% & Fal\% & Und\% 
\\

\hline\hline
\multirow{8}{*}{\rotatebox[origin=c]{90}{Bank }} 
& BM-1
& 30m & \cmark & 11            & 10.00 & 0 & 90.00 
& 4.82s & \cmark & 2820        & 94.23 & 0 & 5.76 
\\ 

& BM-2
& 31m & \cmark & 28            & 16.07 & 0 & 83.93 
& 3.23s & \cmark & 2479        & 93.41 & 0 & 6.58 
\\ 

& BM-3
& 31m & \cmark & 27            & 19.60 & 0 & 81.40 
& 1.21s & \cmark & 1864        & 95.69 & 0 & 4.30 
\\ 

& BM-4
& 35m & \cmark & 4             & 3.72 & 0 & 96.18 
& 71.12s & \cmark & 5135       & 87.03 & 0 & 12.96 
\\ 

& BM-5
& 23m & \cmark & 114           & 77.25 & 0 & 22.75 
& 1.03s & \cmark & 1474        & 96.27 & 0 & 3.72  
\\ 

& BM-6
& 12m & \cmark & 155           & 69.41 & 0 & 30.59 
& 0.44s & \cmark & 1426        & 96.44 & 0 & 3.55 
\\ 

& BM-7
& 30m & \cmark & 57            & 9.41 & 0 & 90.59 
& 12.26s & \cmark & 7017       & 83.65 & 0 & 16.34 
\\ 

& BM-8
& 30m & \cmark & 1             & 0.98 & 0 & 99.02 
& 18.99s & \cmark & 3074       & 90.75 & 0 & 9.24 
\\

\hline
\multirow{5}{*}{\rotatebox[origin=c]{90}{German }} 
& GC-1 
& 32m & \cmark & 22            & 0 & 0 & 100
& 9.73s & \cmark & 31585       & 32.67 & 0 & 67.33
\\ 

& GC-2 
& 33m & \cmark & 6             & 0 & 0 & 100
& 31.72s & \cmark & 32655      & 42.21 & 0 & 57.79
\\

& GC-3 
& 8m & \cmark & 194            & 2.98 & 0 & 97.02
& 6.77s & \cmark & 25963       & 58.44 & 0 & 41.55
\\

& GC-4 
& 4m & \cmark & 2              & 99.00 & 0 & 1.00
& 0.29s & \cmark & 77          & 99.65 & 0 & 0.34
\\

& GC-5
& 30m & \xmark & 0             & 0 & 0 & 100
& 1.24s & \cmark & 9           & 99.80 & 0 & 0.19
\\

\hline
\multirow{12}{*}{\rotatebox[origin=c]{90}{Adult}} 
& AC-1
& 32m & \cmark & 3             & 0.03 & 0 & 99.97 
& 3.23s & \cmark & 6151        & 90.68 & 0 & 9.31  
\\ 

& AC-2
& 31m & \cmark & 9             & 0.01 & 0 & 99.99 
& 30.04s & \cmark & 13008      & 79.93 & 0 & 20.06  
\\ 

& AC-3
& 32m & \cmark & 20            & 0 & 0 & 100 
& 37.12s & \cmark & 60494      & 33.29 & 0 & 66.70  
\\ 

& AC-4
& 36m & \xmark & 0             & 0 & 0 & 100 
& 8m & \cmark & 61324          & 24.79 & 0 & 75.20  
\\ 

& AC-5
& 33m & \xmark & 0             & 0 & 0 & 100 
& 4m & \cmark & 71012          & 19.12 & 0 & 80.87  
\\ 

& AC-6
& 33m & \cmark & 4             & 0.01 & 0 & 99.99 
& 10.20s & \cmark & 31593      & 58.82 & 0 & 41.17 
\\ 

& AC-7
& 30m & \xmark & 0             & 0.01 & 0 & 99.99 
& 4m & \cmark & 25588          & 31.72 & 0 & 68.27  
\\ 

& AC-8
& 30m & \cmark & 39            & 0.03 & 0 & 99.97 
& 11.18s & \cmark & 26179      & 66.50 & 0 & 33.49  
\\ 

& AC-9
& 30m & \cmark & 126           & 0.64 & 0 & 99.36  
& 3.50s & \cmark & 5470        & 91.13 & 0 & 8.86  
\\ 

& AC-10
& 32m & \cmark & 8             & 0.03 & 0 & 99.97  
& 5.01s & \cmark & 9033        & 87.65 & 0 & 12.34  
\\ 

& AC-11
& 30m & \xmark & 0             & 0 & 0 & 100  
& 36.44s & \cmark & 24516      & 58.01 & 0 & 41.98  
\\ 

& AC-12
& 30m & \xmark & 0             & 0.02 & 0 & 99.98  
& 0.91s & \cmark & 8824        & 70.82 & 0 & 29.17  
\\

\hline
\multirow{7}{*}{\rotatebox[origin=c]{90}{Compas}} 
& compas-1
& 17m & \cmark & 2             & 80.00 & 0.32 & 19.68
& 0.01s & \cmark & 17          & 97.27 & 2.72 & 0
\\ 

& compas-2
& 31m & \xmark & 0             & 0 & 0 & 100
& 0.01s & \cmark & 15          & 97.59 & 2.40 & 0
\\ 

& compas-3
& 30m & \xmark & 0             & 0 & 0 & 100
& 0.30s & \cmark & 12          & 98.07 & 1.92 & 0
\\ 

& compas-4
& 30m & \xmark & 0             & 0 & 0 & 100
& 0.01s & \cmark & 14          & 97.75 & 2.24 & 0
\\ 

& compas-5
& T/O & \xmark & 0             & 0 & 0 & 100
& 5.24s & \cmark & 11          & 98.23 & 1.76 & 0
\\ 

& compas-6
& M/O & \xmark & 0             & 0 & 0 & 100
& 9.19s & \cmark & 12          & 98.07 & 1.92 & 0
\\ 

& compas-7
& M/O & \xmark & 0             & 0 & 0 & 100
& 101.25s & \cmark & 15        & 97.59 & 2.40 & 0
\\ 

% \hline
% \multirow{7}{*}{\rotatebox[origin=c]{90}{Salary}} 
% & salary-1
% & 30m & \cmark & 66            & 22.00 & 1.20 & 76.80
% & 0.02s & \cmark & 513         & 90.60 & 9.39 & 0
% \\ 

% & salary-2
% & 30m & \xmark & 0             & 0 & 0 & 100
% & 0.12s & \xmark & 0           & 100 & 0 & 0
% \\

% & salary-3
% & 30m & \xmark & 0             & 0 & 0 & 100
% & 2.69s & \cmark & 428         & 92.16 & 7.83 & 0
% \\

% & salary-4
% & 30m & \xmark & 0             & 0 & 0 & 100
% & 0.14s & \cmark & 574         & 89.48 & 10.51 & 0
% \\

% & salary-5
% & T/O & \xmark & 0             & 0 & 0 & 100
% & 51.75s & \cmark & 271        & 95.03 & 4.96 & 0
% \\ 

% & salary-6
% & M/O & \xmark & 0             & 0 & 0 & 100
% & 2m & \cmark & 471            & 91.37 & 8.62 & 0
% \\ 

% & salary-7
% & M/O & \xmark & 0             & 0 & 0 & 100
% & 18m & \cmark & 61            & 98.88 & 1.11 & 0
% \\ 

\hline
\end{tabular}
\renewcommand{\arraystretch}{1.0}
}

\end{table}

\subsection{Experimental Results}

Table~\ref{table.results} shows the results of our method (\FairQuant{}) in comparison with \Fairify{}\footnote{The order in which \Fairify{} sorts the partitions before running the verification query is random and non-deterministic, so there may be minor difference in the reported counterexamples in the original evaluation and ours.}.
Columns 1-2 show the names of the dataset and the DNN model.  
Columns 3-5 show the statistics reported by \Fairify{}, including the time taken, whether a counterexample is found (\textit{Cex}) and the number of counterexamples found (\textit{\#Cex}). T/O or M/O in Column 3 respectively means that \Fairify{} either spent all 30m or ran out of memory in the network pruning step prior to verifying any input partition.
Columns 6-8 show the percentage of certified, falsified and undecided inputs (\textit{Cer\%, Fal\%, Und\%}).
Columns 9-14 show the corresponding results from \FairQuant{}.

\subsubsection{Results for  RQ 1}

To answer the first research question (RQ 1), i.e., whether our method is more accurate than the current state-of-the-art,  we need to compare the results shown in Columns 4-5 (for \Fairify{}) with the results shown in Columns 10-11 (for \FairQuant{}).  
Specifically, Columns 4 and 10 indicate whether the tool is able to find a counterexample (\cmark) or not (\xmark) within the time limit. 

While our method (\FairQuant{}) found counterexamples for all {32} DNNs, \Fairify{} found counterexamples for only {20} of the {32} models. In addition, it found counterexamples for only {one} of the {7} newly added models.
Moreover, Columns 5 and 11 show that, on models where both tools found counterexamples, the number of counterexamples found by \FairQuant{} is often thousands of times more.  For example, the largest number of counterexamples found by \Fairify{} is 194 (for GC-3) but the large number of counterexamples found by \FairQuant{} is 71,012 (for AC-5).

\subsubsection{Results for RQ 2}

To answer the second research question (RQ 2), i.e., whether our method is more scalable than the current state-of-the-art, we need to compare the running time shown in Column 3 (for \Fairify{}) with the running time shown in Column 9 (for \FairQuant{}).   
While our method (\FairQuant{}) always finished within the time limit of 30 minutes, \Fairify{} {timed out on compas-5 and ran out of memory on compas-6 and compas-7.}
Even on the models where both tools finished, the time taken by \Fairify{} is significantly longer. 

\begin{figure}
    \centering
    \resizebox{\linewidth}{!}{\begin{tikzpicture}

% Start the group plot environment
\begin{groupplot}[
    group style={
        group size=2 by 1, % Number of columns by number of rows
        xlabels at=edge bottom, % Place x axis labels only at the bottom
        ylabels at=edge left, % Place y axis labels only at the left
        horizontal sep=2cm, % Horizontal separation
    },
    ymin=0,
    ybar=2*\pgflinewidth,
    xtick=data,
    x=1.5cm,
    symbolic x coords={
        BM-1,
        GC-1,
        AC-1,
        compas-1,
        % salary-1,
    },
]

% First plot (Runtime)
\nextgroupplot[
    ylabel = {Runtime (m)},
]

\addplot[style={red,fill=lightred,mark=none}]
    coordinates {
    (BM-1,30)
    (GC-1,32)
    (AC-1,32)
    (compas-1,17)
    % (salary-1,30)
    };

\addplot[style={blue,fill=lightblue,mark=none}]
     coordinates {
     (BM-1,0.08)
     (GC-1,0.16)
     (AC-1,0.05)
     (compas-1,0.0001)
     % (salary-1,0.0003)
     };

% Second plot (Undecided)
\nextgroupplot[
    ylabel = {Undecided (\%)},
]

\addplot[style={red,fill=lightred,mark=none}]
    coordinates {
    (BM-1,90.00)
    (GC-1,100.00)
    (AC-1,99.97)
    (compas-1,19.68)
    % (salary-1,76.80)
    };

\addplot[style={blue,fill=lightblue,mark=none}]
     coordinates {
     (BM-1,5.76)
     (GC-1,67.33)
     (AC-1,9.31)
     (compas-1,0.00)
     % (salary-1,0.00)
     };

\end{groupplot}

\end{tikzpicture}}
    \caption{Comparing the runtime overhead (left) and accuracy (right) of \Fairify{}~\cite{biswas_fairify_2023}, in red, and \FairQuant{} (new), in blue.}
    \label{fig:graph}
\end{figure}
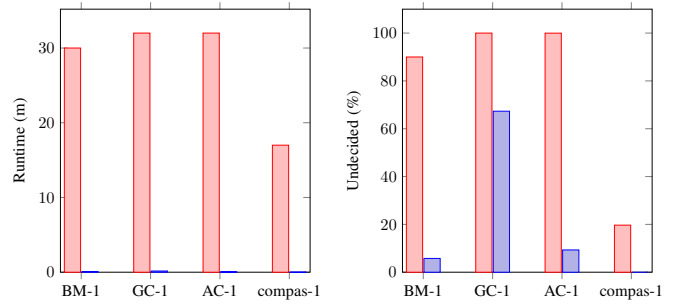

To illustrate the scalability advantage of our method, we took a subset of the models for which both \FairQuant{} and \Fairify{} finished, and plot the running time in a bar chart, shown on the left side of Fig.~\ref{fig:graph}.  Here, the red bars represent the time taken by \Fairify{} and the blue bars represent the time taken by \FairQuant{}.  The results show that \FairQuant{} is many orders of magnitude faster, and can certify DNN models that are well beyond the reach of \Fairify{}.

\subsubsection{Results for RQ 3}

To answer the third research question (RQ 3), i.e., whether our method is more effective in providing feedback to the user, we need to compare the results in Columns 6-8  (for \Fairify{}) with the results in Columns 12-14 (for \FairQuant{}), which show the certified, falsified, and undecided percentages. 
Since \Fairify{} was not designed to quantitatively measure the degree of fairness, it did poorly in almost all cases. Except for a few DNN models for Bank, GC-4, and compas-1, its certified percentages are either 0 or close to 0, and its undecided percentages are almost 100\%.  It means that, for the vast majority of individuals in the input domain, whether they are treated by the DNN model fairly or not remains undecided. 

In contrast, the certified percentages reported by our method (\FairQuant{}) are significantly higher. For the models trained using the Compas dataset, in particular, the certified percentages are around or above 90\%, and more importantly, the undecided percentages are always 0.  It means that \FairQuant{} has partitioned the input domain in such a way that each partition is either certified as being fair, or falsified as being unfair. Even on the subset of DNN models where some inputs remain undecided by \FairQuant{}, the undecided percentages reported by our method are significantly lower than \Fairify{}, as shown on the right side of Fig.~\ref{fig:graph}.

\subsection{Summary}

The results show that our method is more accurate and more scalable than the current state-of-the-art techniques for \emph{qualitative} certification. In addition, our method is able to formally \emph{quantify} the degree of fairness, which is a capability that existing methods do not have. 

For some DNN models, \FairQuant{} still has a significant percentage of inputs left undecided. This is because we set \textit{min\_check\_depth=15} and \textit{max\_refinement\_depth=20} for all benchmarks.  Thus, as soon as \FairQuant{} reaches the refinement depth 15 (see the refinement tree shown in Fig.~\ref{fig:tree}) and finds a counterexample in the input partition, it will stop refining further; at that moment, all inputs in the partition are treated \emph{conservatively} as undecided.

In general, a smaller refinement depth allows \FairQuant{} to terminate quickly. During our experiments, \FairQuant{} terminated after 0.29s and 1.24s for GC-4 and GC-5, respectively, compared to the 4 minutes and 30 minutes taken by \Fairify{} and yet returned better results.  
In fact, for GC-5, \Fairify{} spent 30 minutes but failed to find any counterexample. 
If we increase the refinement depth, by increasing the two threshold values of \FairQuant{}, its quantification results will get even better.

\section{Related Work}
\label{sec:related}

Our method is the first scalable method for certifying and quantifying individual fairness of a deep neural network, and it outperforms the most closely related prior work, \Fairify{}~\cite{biswas_fairify_2023}.
% which is the only tool currently available for certifying individual fairness of a DNN. 
%
To the best of our knowledge, no other methods can match the accuracy, scalability, and functionality of our method.

Our method differs from existing techniques for verifying individual fairness properties for neural networks.
Libra~\cite{urban_perfectly_2020, mazzucato_reduced_2021} uses abstract domains to perform verification but is limited in scalability due to the expensive pre-analysis, where a network of 20 hidden nodes takes several hours even with leveraging multiple CPUs.
DeepGemini~\cite{xie2023deepgemini}, which outperforms Libra, is built on top of Marabou~\cite{katz2019marabou}, a SMT-based neural network verification tool, and thus shares the same limitations of \Fairify{}. Furthermore, it only evaluates on networks with up to around 250 hidden neurons.
Other works have tackled neural network verification of different definitions of individual fairness.
Benussi et al.~\cite{benussi2022individual} and Khedr et al.~\cite{khedr2023certifair} proposed different methods to certify a definition of global individual fairness proposed in~\cite{john2020verifying}.
Ruoss et al.~\cite{ruoss2020learning} verify a type of a \emph{local} individual fairness property that is similar to local robustness, given an input $x$ and a small constant $\varepsilon$ for perturbation. This is different from a \emph{global} perspective we have discussed so far in this paper.

Group fairness is yet another type of fairness property, which can be verified using probabilistic techniques~\cite{albarghouthi2017fairsquare, bastani2019probabilistic, feldman_certifying_2015}. 
The difference between individual fairness and group fairness is that, while individual fairness requires \emph{similar individuals} to be treated similarly, group fairness requires \emph{similar demographic groups} to be treated similarly.

There are other prior works related to fairness verification of other types of machine learning models~\cite{john2020verifying, yannan_fair_knn, kusner2017counterfactual}, but they are not applicable to deep neural networks.
Testing techniques can quickly detect fairness violations in machine learning models including neural networks~\cite{aggarwal_testing, galhotra_fairness_2017, angell_themis_2018, udeshi_testing,zhang_ADF_2020, zheng2022neuronfair}, but it does not provide formal guarantee that is important for certain applications. 
There are also techniques for improving fairness of machine learning models~\cite{yurochkin2020training, jingbo_fair_dt,gao_fairneuron_2022,ranzato_fair_2021,lahoti2020fairness,hashimoto2018fairness,hardt2016equality}, which are orthogonal to our method that focuses on certifying and quantifying fairness of existing DNN models.

At a high level, our method is related to the large number of robustness verifiers for deep neural networks based on interval analysis~\cite{wang_formal_2018, shiqi2018neurify, singh_abstract_2019, yang2021improving}, SMT solving~\cite{ehlers2017formal, katz2017reluplex, katz2019marabou, huang2017safety}, and mixed-integer linear programming~\cite{bastani_measuring_2016, bunel2020bnb}.
While these verifiers can decide if a model is robust against adversarial perturbation, they cannot directly certify individual fairness, as explained earlier in Section~\ref{sec:motivation}.
Other neural network verifiers that deal with differential~\cite{paulsen_reludiff_2020, paulsen2020neurodiff, mohammadinejad2021diffrnn} or equivalence verification~\cite{eleftheriadis2022neural} are also different, since they evaluate over two networks instead of one network.

\section{Conclusion}
\label{sec:conclusion}

We have presented \FairQuant{}, a scalable method for certifying and quantifying individual fairness of a deep neural network over the entire input domain.  It relies on sound abstraction during symbolic forward analysis to improve scalability, and iterative refinement based on backward analysis to improve accuracy. In addition to certifying fairness, it is able to quantify the degree of fairness by computing the percentages of inputs whose  classification outputs can be certified as fair or falsified as unfair. 
We have evaluated the method on a large number of DNN models trained using four popular fairness research datasets. The experimental results show that the method significantly outperforms state-of-the-art techniques in terms of both accuracy and scalability, as well as the ability to quantify the degree of fairness.

\section*{Acknowledgments}

This research was supported in part by the U.S.\ National
Science Foundation (NSF) under grant CCF-2220345. 
We thank the anonymous reviewers for their constructive feedback.

%\clearpage
%\newpage
\nocite{*}

\bibliographystyle{ieeetr}
\bibliography{main}

\end{document}